\documentclass[10pt,twocolumn,letterpaper]{article}

\newcommand\blfootnote[1]{%
  \begingroup
  \renewcommand\thefootnote{}\footnote{#1}%
  \addtocounter{footnote}{-1}%
  \endgroup
}

\usepackage{cvpr}
\usepackage{times}
\usepackage{epsfig}
\usepackage{graphicx}
\usepackage{amsmath}
\usepackage{amssymb}
\usepackage{subcaption}

\usepackage[pagebackref=true,breaklinks=true,letterpaper=true,colorlinks,bookmarks=false]{hyperref}

\cvprfinalcopy 

\def\cvprPaperID{4131} 

\ifcvprfinal\fi
\begin{document}

\title{PointNetLK: Robust \& Efficient Point Cloud Registration using PointNet}

\author{Yasuhiro Aoki\textsuperscript{1,2}* \qquad Hunter Goforth\textsuperscript{1}* \qquad Rangaprasad Arun Srivatsan\textsuperscript{1} \qquad Simon Lucey\textsuperscript{1,3} \\
\textsuperscript{1}Carnegie Mellon University \qquad \textsuperscript{2}Fujitsu Laboratories Ltd. \qquad \textsuperscript{3}Argo AI \\
{ \tt\small aoki-yasuhiro@fujitsu.com} \qquad \tt\small \{hgoforth,arangapr,slucey\}@cs.cmu.edu}

\maketitle
\thispagestyle{empty}

\begin{abstract}
PointNet has revolutionized how we think about representing point clouds. For classification and segmentation tasks, the approach and its subsequent extensions are state-of-the-art. To date, the successful application of PointNet to point cloud registration has remained elusive. In this paper we argue that PointNet itself can be thought of as a learnable ``imaging'' function. As a consequence, classical vision algorithms for image alignment can be applied on the problem -- namely the Lucas \& Kanade (LK) algorithm. Our central innovations stem from: (i) how to modify the LK algorithm to accommodate the PointNet imaging function, and (ii) unrolling PointNet and the LK algorithm into a single trainable recurrent deep neural network. We describe the architecture, and compare its performance against state-of-the-art in common registration scenarios. The architecture offers some remarkable properties including: generalization across shape categories and computational efficiency -- opening up new paths of exploration for the application of deep learning to point cloud registration. Code and videos are available at \url{https://github.com/hmgoforth/PointNetLK}. \blfootnote{* equal contribution.}
\end{abstract}


\section{Introduction}
Point clouds are inherently unstructured with sample and order permutation ambiguities. This lack of structure makes them problematic for use in modern deep learning architectures. PointNet~\cite{qi2017pointnet} has been revolutionary from this perspective, as it offers a learnable structured representation for point clouds. One can think of this process as a kind of ``imaging'' -- producing a fixed dimensional output irrespective of the number of samples or ordering of points. This innovation has produced a number of new extensions and variants~\cite{qi2017pointnet++, shen2017neighbors,wang2018dynamic} that are now state-of-the-art in object classification and segmentation on point clouds.

\begin{figure}[h!]
\centering
\includegraphics[width=0.45\textwidth]{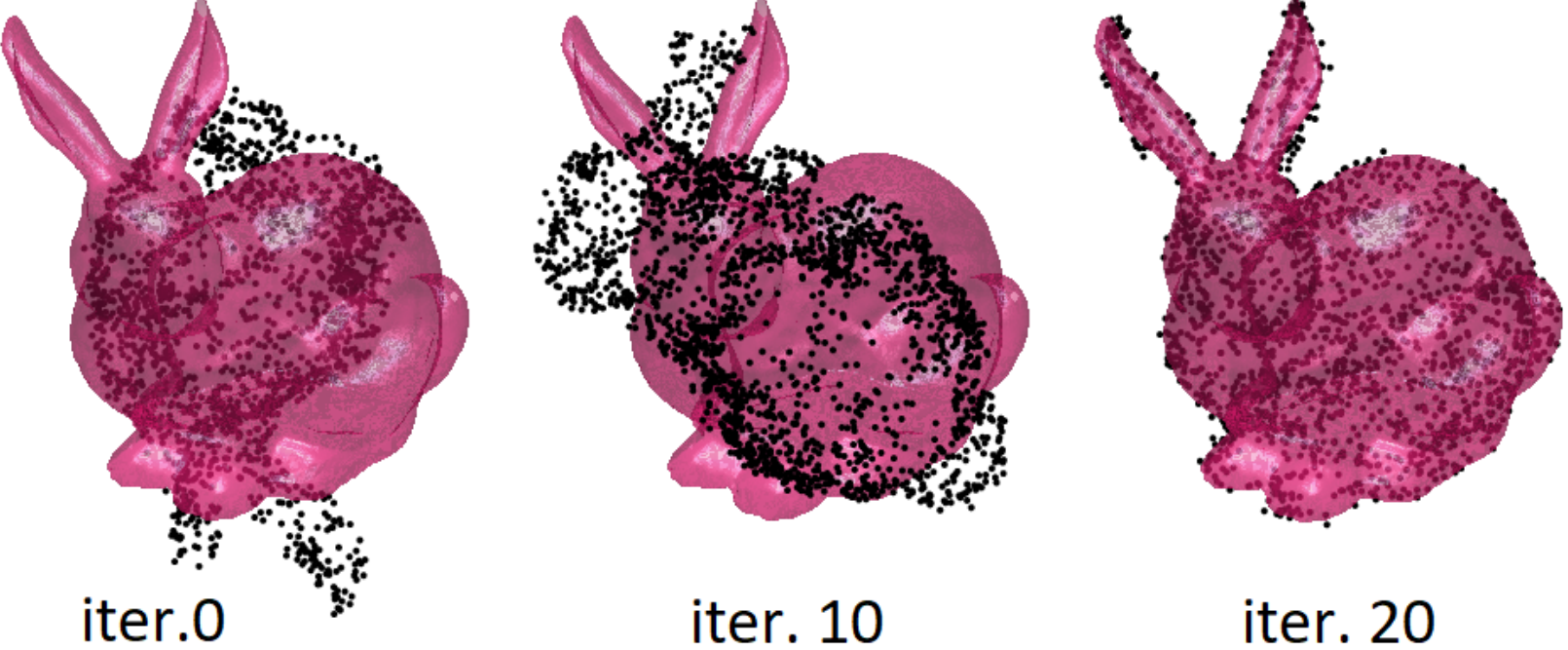}
\includegraphics[width=0.45\textwidth]{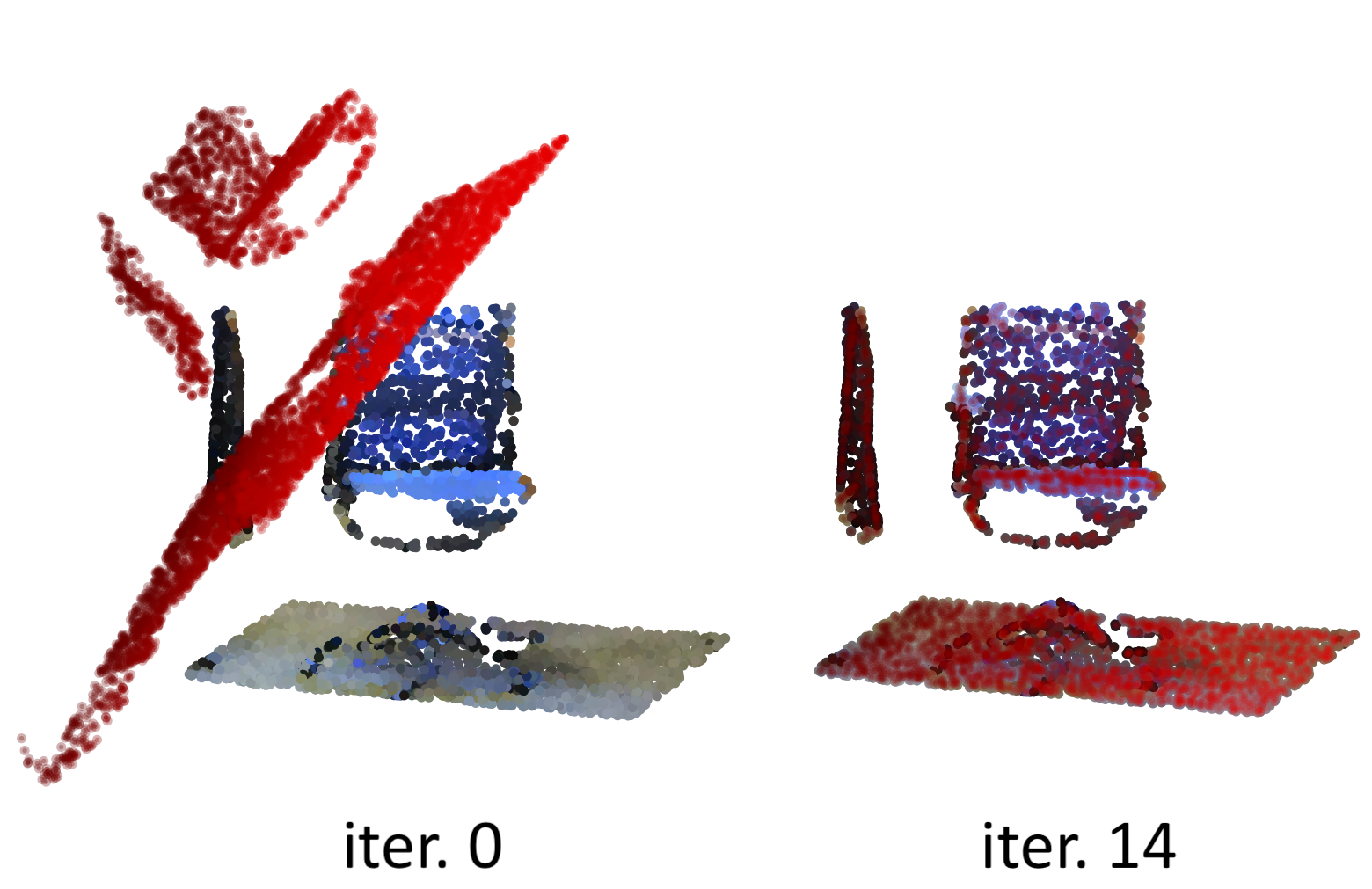}
\caption{Point cloud registration of (Top) Stanford bunny~\cite{turk2005stanford} and (Bottom) raw indoor scan from S3DIS~\cite{Armeni_2016_CVPR} with PointNetLK. Refer to Sec.~\ref{train_test_different_category} and  Sec.~\ref{partially_visible_data} for more details. As the iterations progress, PointNetLK is able to successfully register the source points to the template model, even though it was not trained on these shapes. We include Bunny surface rendering for the sake of visualization.}
\label{fig_bunny}
\end{figure}

The utility of PointNet for the task of point cloud registration, however, has remained somewhat elusive. In this paper we want to explore further the notion of interpreting the PointNet representation as an imaging function -- a direct benefit of which could be the application of image alignment approaches to the problem of point cloud registration. In particular we want to utilize the classical Lucas \& Kanade (LK) algorithm~\cite{lucas1981iterative}. This connection is motivated by a recent innovation~\cite{wang2018deep} that has demonstrated state-of-the-art 2D photometric object tracking performance by reinterpreting the LK algorithm as a recurrent neural network.

The LK algorithm, however, cannot be naively applied to the PointNet representation. This is due to the LK algorithm's dependence on gradient estimates, which are estimated in practice through convolution. Within a 2D photometric image, or a 3D volumetric image, each element of the representation (i.e. pixel or voxel) has a known local dependency between its neighbors, which can be expressed as 2D- and 3D- grids respectively -- from which convolution can be defined. It is also well understood that this dependency does not have to take the form or a $N$D-grid, with the notion of ``graph'' convolution~\cite{wang2018dynamic} also being explored. PointNet representations have no such local dependency making the estimation of spatial gradients through convolution ill posed. 

\paragraph{Contributions:}
We propose a modification to the LK algorithm which circumvents the need for convolution on the PointNet representation. We then demonstrate how this modified LK form can be unrolled as a recurrent neural network and integrated within the PointNet framework -- this unified network shall be referred to herein as PointNetLK. Unlike many variants of iterative closest point (ICP), our approach requires no costly computation of point correspondences~\cite{rusinkiewicz2001},  which gives rise to substantial advantages in terms of accuracy, robustness to initialization and computational efficiency. PointNetLK exhibits remarkable generalization to unseen object and shape variations, as shown in Fig.~\ref{fig_bunny}. This generalization performance can be attributed to the explicit encoding of the alignment process within the network architecture. As a consequence, the network only needs to learn the PointNet representation rather than the task of alignment. Finally, our approach is fully differentiable, unlike most registration approaches in literature, hence allowing for an easy integration with larger DNN systems.  An added computational benefit is that our approach can be run directly on GPU as part of a larger neural-network pipeline, unlike most of the comparisons which require a method like ICP or its variants to be run on CPU.

\section{Related Work}
\paragraph{PointNet:}
PointNet~\cite{qi2017pointnet} is the first work to propose the use of DNN with raw point clouds as input, for the purposes of classification and segmentation. The architecture achieves state of the art performance on this task despite its simplicity, and provides interesting theoretical insight into processing raw point clouds. PointNet++ was proposed as an improvement over the PointNet, by hierarchically aggregating features in local point sets~\cite{qi2017pointnet++}. Another variant considers aggregates features of nearby points~\cite{shen2017neighbors}. Wang~\etal~\cite{wang2018dynamic} use a local neighborhood graph and convolution-like operations on the edges connecting
neighboring pairs of points. 

\paragraph{ICP and variants:}
Besl and McKay~\cite{besl1992method} introduced the iterative closest point (ICP), which is a popular approach for registration, by iteratively estimating point correspondence and performing a least squares optimization. 
Several variants of the ICP have been developed~(see \cite{rusinkiewicz2001} for a review)  that incorporate sensor uncertainties~\cite{segal2009generalized,arun2018probabilistic}, are robust to outliers~\cite{bouaziz2013sparse}, use different optimizers~\cite{fitzgibbon2003robust}, etc. ICP and its variants, however, have a few fundamental drawbacks, namely: (1) explicit estimation of closest point correspondences, which results in the complexity scaling quadratically with the number of points, (2) sensitive to initialization, and (3)  nontrivial to integrate them to deep learning framework due to issues of differentiability.

\paragraph{Globally optimal registration:} Since ICP and most of its variants are sensitive to initial perturbation in alignment, they only produce locally optimal estimates. Yang~\etal~\cite{Yang13} developed Go-ICP, a branch and bound-based optimization approach to obtain globally optimal pose. More recently convex relaxation has been used for global pose estimation using Riemannian optimization~\cite{rosen2016certifiably}, semi-definite programming~\cite{horowitz2014convex,maron2016point} and mixed integer programming~\cite{izatt2017}. A major drawback of the above methods is the large computation time, rendering them unsuitable for real time applications.

\paragraph{Interest point methods:}
There are works in literature that estimate interest points to help with registration. For instance, scale invariant curvature descriptors~\cite{gelfand2005robust}, oriented descriptors~\cite{glover2012}, extended Gaussian images~\cite{makadia2006fully}, fast point feature histograms~\cite{rusu2009fast}, color intensity-based descriptors~\cite{godin1994three}, global point signatures~\cite{chua1997point}, heat kernels~\cite{ovsjanikov2010one}, etc. While interest points have the potential to improve the computationally speed of the registration approaches, they do not generalize to all applications~\cite{guo20143d}.

\paragraph{Hand-crafted representations:} 
The discriminative optimization (DO) work of Vongkulbhisal~\etal~\cite{vongkulbhisal2017discriminative} uses a hand-crafted feature vector and learns a set of maps, to estimate a good initial alignment. The alignment is later refined using an ICP. The drawback of this approach is that the features and maps are specific to each object and do not generalize. More recently they developed inverse composition discriminative optimization (ICDO), which generalizes over unseen object shapes. ICDO unfortunately has a complexity which is quadratic in the number of points, making it difficult to use in several real world scenarios. Another issue with ICDO is that both the features and alignment maps are learned, which can result in a compromise on the generalizability of the approach.

\paragraph{Alternate representations:}
Voxelization is a method to discretize the space and convert a point clouds to a structured grid. Several methods have been developed that use DNNs over voxels~\cite{maturana2015voxnet, wu20153d}. Major drawbacks of these approaches include computation time and memory requirements. Another popular representation is depth image or range image, which represents the point cloud as a collection of 2D views, which are easily obtained by commercial structured light sensors. Typically convolution operations are performed on each view and the resulting features are  aggregated~\cite{su2015multi}. Some works also combine voxel data with multi-view data~\cite{qi2016volumetric,balntas2017pose}. There are several works that directly estimate 3D pose from photometric images. For instance,~\cite{su2015render, kendall2015posenet, massa2016crafting, xiang2016objectnet3d, mousavian20173d}, directly regress over the Euler angles of object orientations from cropped object images. On the other hand, in applications such as robotic manipulation, pose is often decoupled into rotation and translation components and each is inferred independently~\cite{su2015render, tekin2017real, kehl2017ssd, xiang2017posecnn, rad2017bb8, li2018unified}.

\section{PointNetLK}

\begin{figure*}
\begin{center}
\includegraphics[width=18cm,height=6cm,trim={2cm 7cm 3cm 1cm},clip]{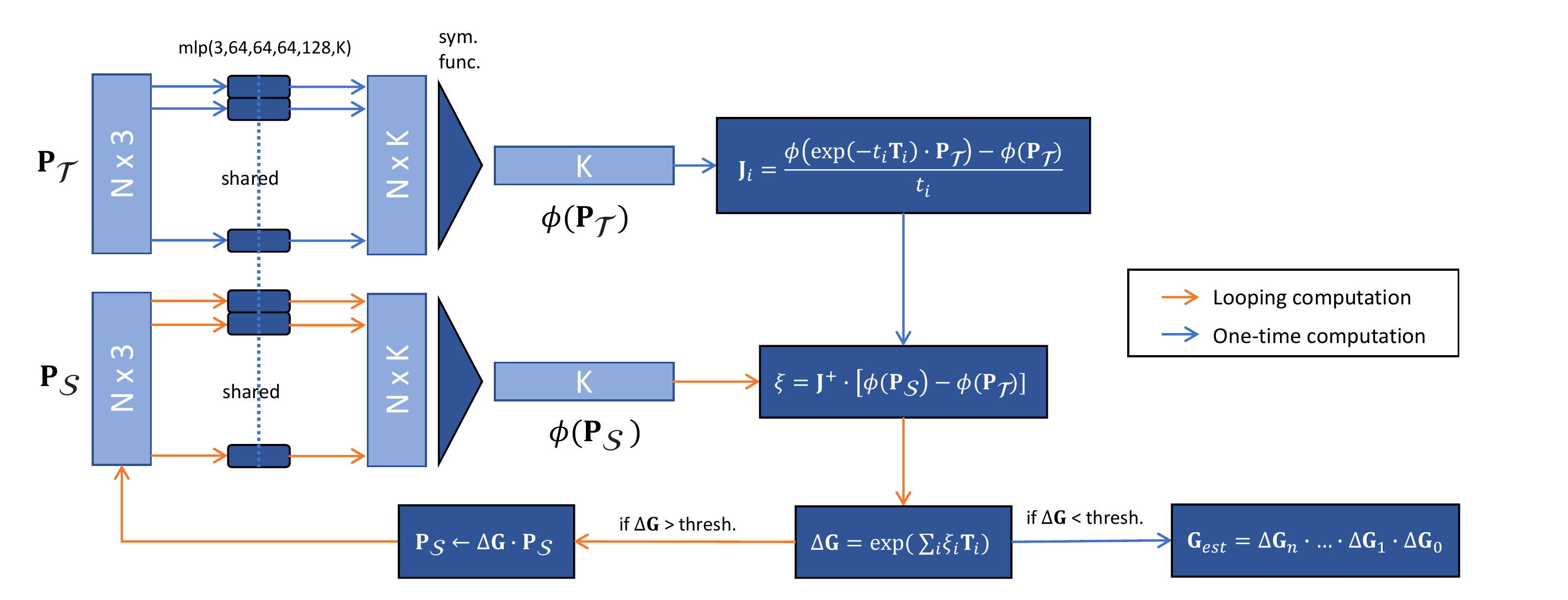}
\end{center}
   \caption{Point cloud inputs source $\textbf{P}_\mathcal{S}$ and template $\textbf{P}_\mathcal{T}$ are passed through a shared MLP, and a symmetric pooling function, to compute the global feature vectors $\phi(\textbf{P}_\mathcal{S})$ and $\phi(\textbf{P}_\mathcal{T})$. The Jacobian $\textbf{J}$ is computed once using $\phi(\textbf{P}_\mathcal{T})$. The optimal twist parameters are found, which are used to incrementally update the pose of $\textbf{P}_\mathcal{S}$, and then the global feature vector $\phi(\textbf{P}_\mathcal{S})$ is recomputed. During training, a loss function is used which is based on the difference in the estimated rigid transform and the ground truth transform.}
\label{fig_block}
\end{figure*}
In Section \ref{method_overview} we introduce notation and mathematics for PointNetLK. In Section \ref{method_derivation} we provide a derivation of the optimization on PointNet feature vectors used for point cloud alignment. In Section \ref{method_training} we describe aspects of training for our model, including loss functions and possible symmetric operators. 
\paragraph{Notation:} We denote matrices with uppercase bold such as $\textbf{M}$, constants as uppercase italic such as $C$, and scalar variables with lowercase italic such as $s$.

\subsection{Overview} \label{method_overview}
Let $\phi$ denote the PointNet function, 
\mbox{$\phi:\mathbb{R}^{3\times N}\rightarrow \mathbb{R}^K$}, such that for an input point cloud $\textbf{P} \in \mathbb{R}^{3 \times N}$,  $\phi(\textbf{P})$ produces a $K$-dimensional vector descriptor. The function $\phi$ applies a Multi-Layer Perceptron (MLP) to each 3D point in $\textbf{P}$, such that the final output dimension of each point is $K$. Then a symmetric pooling function, such as maximum or average, is applied, resulting in the $K$-dimensional global descriptor.

We formulate an optimization as follows. Let $\textbf{P}_\mathcal{T}$, $\textbf{P}_\mathcal{S}$ be template and source point clouds respectively. We will seek to find the rigid-body transform $\textbf{G} \in SE(3)$ which best aligns source $\textbf{P}_\mathcal{S}$ to template $\textbf{P}_\mathcal{T}$. The transform $\textbf{G}$ will be represented by an exponential map as follows:

\begin{equation} \label{eqn_G}
\textbf{G} = \exp \left(\sum_i \xi_i \textbf{T}_i \right) \qquad \boldsymbol{\xi} = (\xi_1, \xi_2, ..., \xi_6)^T,
\end{equation}
where $\textbf{T}_i$ are the generators of the exponential map with twist parameters $\boldsymbol{\xi}\in \mathbb{R}^6$. The 3D point cloud alignment problem can then be described as finding $\textbf{G}$ such that $\phi(\textbf{P}_\mathcal{T}) = \phi(\textbf{G} \cdot \textbf{P}_\mathcal{S})$, where we use the shorthand $(\cdot)$ to denote transformation of $\textbf{P}_\mathcal{S}$ by rigid transform $\textbf{G}$. This equation is analogous to the quantity being optimized in the classical LK algorithm for 2D images, where the source image is warped such that the pixel intensity differences between the warped source and template are minimized.  It is worth noting that we do not include the T-net in our PointNet architecture, since its purpose was to transform the input point cloud in order to increase classification accuracy~\cite{qi2017pointnet}.   However,  we instead use the LK layer to estimate the alignment, and the T-net is unnecessary.

Another key idea that we can borrow from the LK algorithm is the Inverse Compositional (IC) formulation~\cite{baker2004lucas}. The IC formulation is necessitated by the fact that the traditional LK algorithm has a high computational cost for each iteration of the optimization. This cost comes from the re-computation of an image Jacobian on the warped source image, at each step of the optimization. The insight of the IC formulation is to reverse the role of the template and source: at each iteration, we will solve for the incremental warp update to the template instead of the source, and then apply the \emph{inverse} of this incremental warp to the source. By doing this, the Jacobian computation is performed for the template instead of the source and happens only once before the optimization begins. This fact will be more clearly seen in the following derivation of the warp update.

\subsection{Derivation} \label{method_derivation}
Restating the objective, we seek to find $\textbf{G}$ such that $\phi(\textbf{P}_\mathcal{T}) = \phi(\textbf{G} \cdot \textbf{P}_\mathcal{S})$. To do this, we will derive an iterative optimization solution. 

With the IC formulation in mind, we take an inverse form for the objective:

\begin{equation} \label{eqn_inv_form}
\phi(\textbf{P}_\mathcal{S}) = \phi(\textbf{G}^{-1} \cdot \textbf{P}_\mathcal{T})
\end{equation}

The next step is to linearize the right-hand side of (\ref{eqn_inv_form}):

\begin{equation} \label{eqn_taylor}
\phi(\textbf{P}_\mathcal{S}) = \phi(\textbf{P}_\mathcal{T}) + \frac{\partial}{\partial \boldsymbol{\xi}} \left[ \phi(\textbf{G}^{-1} \cdot \textbf{P}_\mathcal{T}) \right] \boldsymbol{\xi}
\end{equation}

Where we define $\textbf{G}^{-1} = \exp (-\sum_i \xi_i \textbf{T}_i )$.

\paragraph{Canonical LK:} 

We will denote the Jacobian \mbox{$\textbf{J} = \frac{\partial}{\partial \boldsymbol{\xi}} \left[ \phi(\textbf{G}^{-1} \cdot \textbf{P}_\mathcal{T}) \right]$}, where $\textbf{J}\in \mathbb{R}^{K \times 6}$ matrix. At this point, computing $\textbf{J}$ would seem to require an analytical representation of the gradient for the PointNet function with respect to the twist parameters of $\textbf{G}$. This analytical gradient would be difficult to compute and quite costly. The approach taken in the classical LK algorithm for $N$D images is to split the Jacobian using the chain rule, into two partial terms: an image gradient in the $N$D image directions, and an analytical warp Jacobian~\cite{baker2004lucas}. However, in our case this approach will not work either, since there is no graph or other convolutional structure which would allow taking gradients in $x$, $y$ and $z$ for our 3D registration case.

\paragraph{Modified LK:}
Motivated by these challenges, we instead opt to compute $\textbf{J}$ using a stochastic gradient approach. Specifically, each column $\textbf{J}_i$ of the Jacobian can be approximated through a finite difference gradient computed as

\begin{equation} \label{eqn_jacob}
\textbf{J}_i = \frac{\phi(\exp(-t_i \textbf{T}_i) \cdot \textbf{P}_\mathcal{T}) - \phi(\textbf{P}_\mathcal{T})}{t_i}
\end{equation}

Where $t_i$ are infinitesimal perturbations of the twist parameters $\boldsymbol{\xi}$. This approach to computing $\textbf{J}$ is what allows the application of the computationally efficient inverse compositional LK algorithm to the problem of point cloud registration using PointNet features. Note that $\textbf{J}$ is computed only once, for the template point cloud, and does not need to be recomputed as the source point cloud is warped during iterative alignment.

For each column $\textbf{J}_i$ of the Jacobian, only the $i^{th}$ twist parameter has a non-zero value $t_i$. Theoretically, $t_i$ should be infinitesimal so that $\textbf{J}$ is equal to an analytical derivative. In practice, we find empirically that setting $t_i$ to some small fixed value over all iterations yields the best result.

We can now solve for $\boldsymbol{\xi}$ in (\ref{eqn_taylor}) as

\begin{equation} \label{eqn_twist}
\boldsymbol{\xi} = \textbf{J}^{+} \left[\phi(\textbf{P}_\mathcal{S}) - \phi(\textbf{P}_\mathcal{T})\right]
\end{equation}

Where $\textbf{J}^{+}$ is a Moore-Penrose inverse of $\textbf{J}$. 

In summary, our iterative algorithm consists of a looping computation of the optimal twist parameters using (\ref{eqn_twist}), and then updating the source point cloud $\textbf{P}_\mathcal{S}$ as

\begin{equation} \label{eqn_update}
\textbf{P}_\mathcal{S} \leftarrow \Delta \textbf{G} \cdot \textbf{P}_\mathcal{S} \qquad \Delta \textbf{G} = \exp \left( \sum_i \xi_i \textbf{T}_i \right)
\end{equation}

The final estimate $\textbf{G}_{est}$ is then the composition of all incremental estimates computed during the iterative loop:

\begin{equation} \label{eqn_G_compose}
\textbf{G}_{est} = \Delta \textbf{G}_n \cdot ... \cdot \Delta \textbf{G}_1 \cdot \Delta \textbf{G}_0 
\end{equation}

The stopping criterion for iterations is based on a minimum threshold for $\Delta \textbf{G}$. A graphical representation of our model is shown in Fig. \ref{fig_block}.

\subsection{Training} \label{method_training}

\paragraph{Loss function:} The loss function for training should be targeted at minimizing the difference between the estimated transform $\textbf{G}_{est}$ and the ground truth transform $\textbf{G}_{gt}$. This could be expressed as the Mean Square Error (MSE) between the twist parameters $\boldsymbol{\xi}_{est}$ and $\boldsymbol{\xi}_{gt}$. Instead, we use

\begin{equation} \label{loss2}
||(\textbf{G}_{est})^{-1} \cdot \textbf{G}_{gt} - \textbf{I}_4||_F,
\end{equation}

which is more computationally efficient to compute as it does not require matrix logarithm operation during training, and follows in a straightforward way from the representation of $\textbf{G}_{est}, \textbf{G}_{gt} \in SE(3)$.

\paragraph{Symmetric pooling operator:} 
In PointNet, the MLP operation is followed by a symmetric pooling function such as maximum or average pooling, to facilitate point-order permutation invariance (see Fig. \ref{fig_block}). In Section \ref{experiments}, we show results using either max or average pooling and make observations about which operator may be more suitable given different scenarios. Particularly, we hypothesize that average pooling would have an advantage over max pooling on the case of noisy point cloud data, which is confirmed in our experiments.

\begin{figure}
\centering
\includegraphics[width=0.4\textwidth]{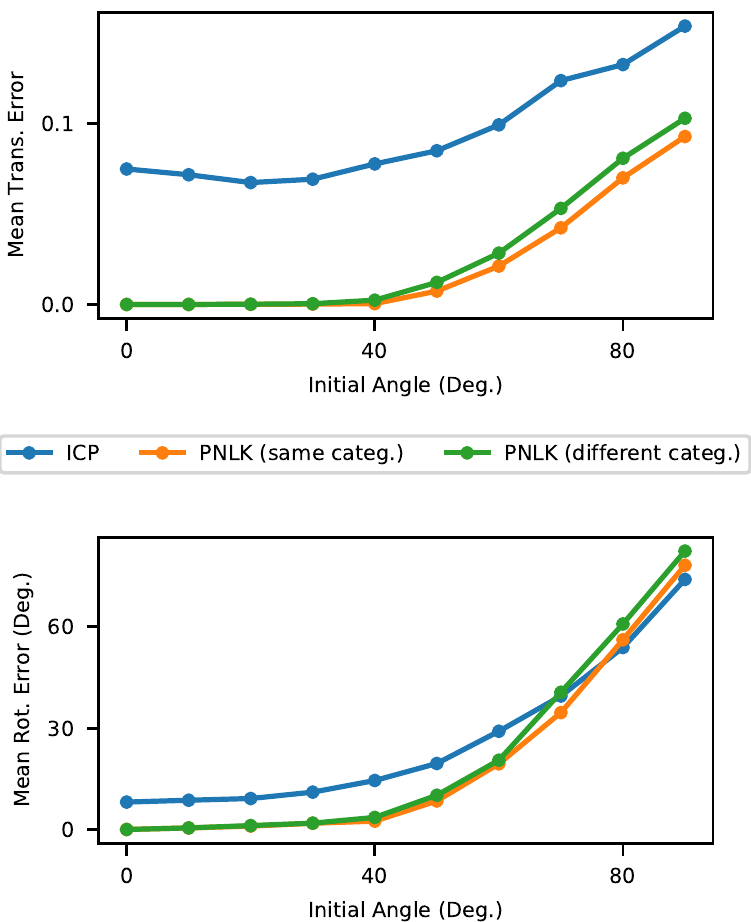}
\caption{Results for Section \ref{train_test_same_object_category} and \ref{train_test_different_category}. PointNetLK achieves remarkable alignment results on categories seen during training (PNLK same category), as well as those unseen during training (PNLK different category). Results are reported for 10 iterations of both PointNetLK and ICP, showcasing also the ability of PointNetLK to align quickly in fewer iterations.}
\label{fig_ex1_ex2}
\end{figure}

\begin{figure}
\centering 
\includegraphics[width=0.4\textwidth]{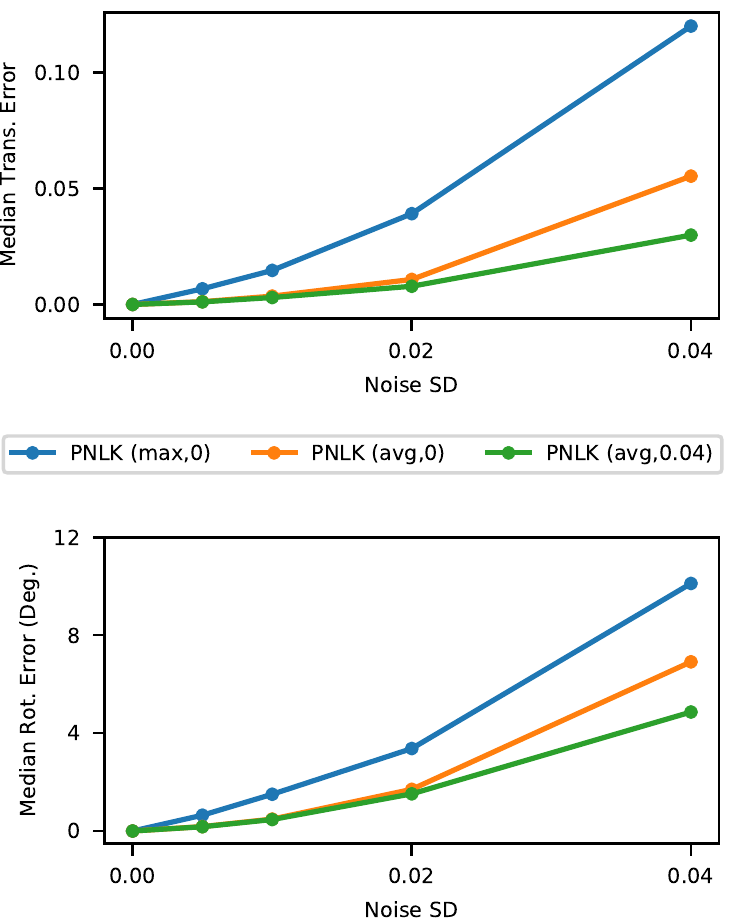}
\caption{Results for Section \ref{gaussian_noise}. We compare PointNetLK trained on zero-noise data with max pool, trained on zero-noise data with avg. pool, and trained on noisy (SD=0.04) data using avg. pool. The results support our hypothesis that avg. pooling is important in order to account for noise in data.}
\label{fig_ex3}
\end{figure}

\section{Experiments} \label{experiments}
We experiment with various combinations of training data, test data, and symmetric operators. We compare with ICP~\cite{besl1992method} as a baseline at test time. We have used ModelNet40~\cite{wu20153d}, a dataset containing CAD models for 40 object categories, for experiments unless otherwise noted.

\begin{figure*}[h!]
\centering
   \includegraphics[width=0.325\textwidth]{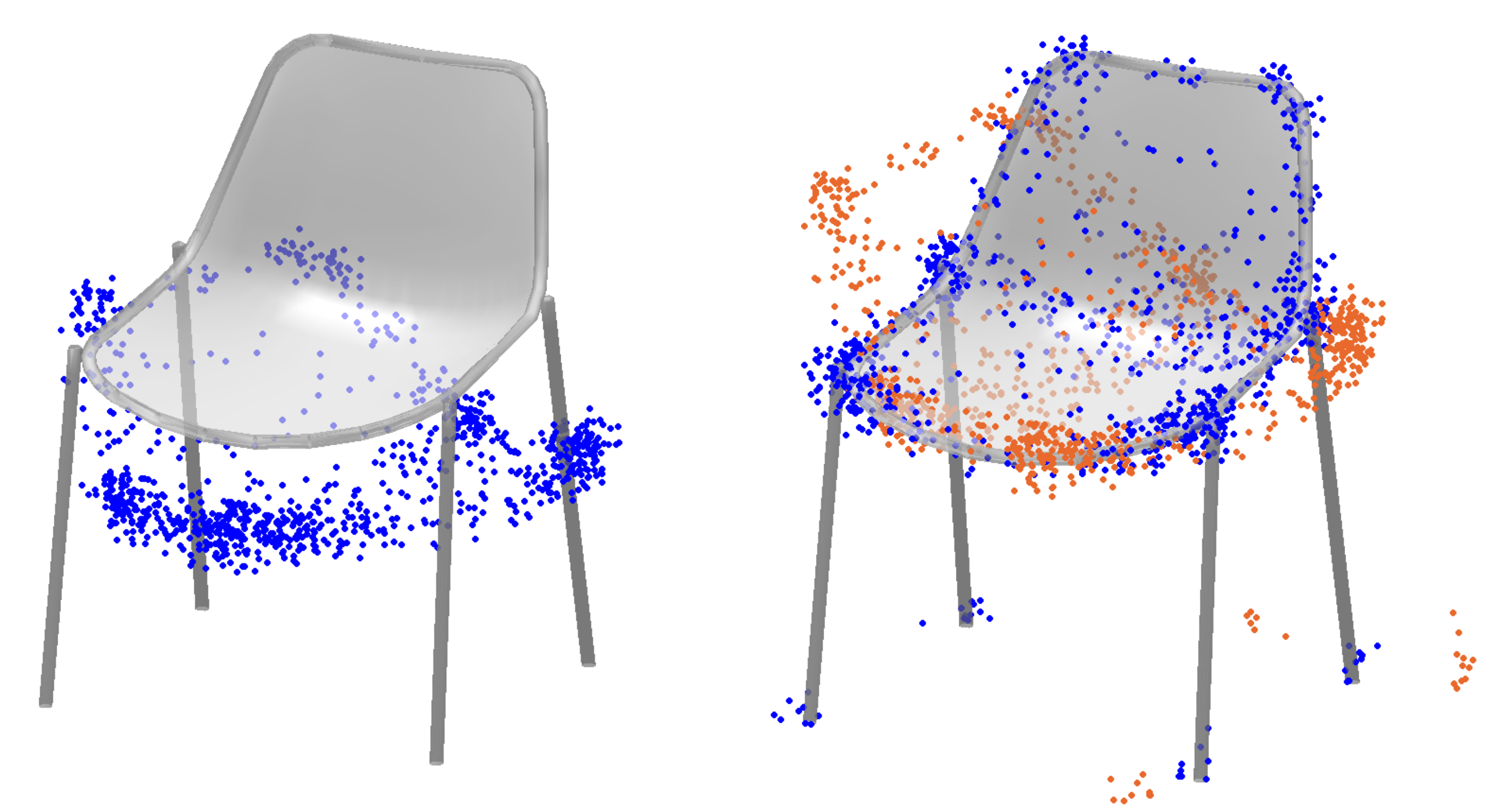}\hspace{20mm}
  \includegraphics[width=0.325\textwidth]{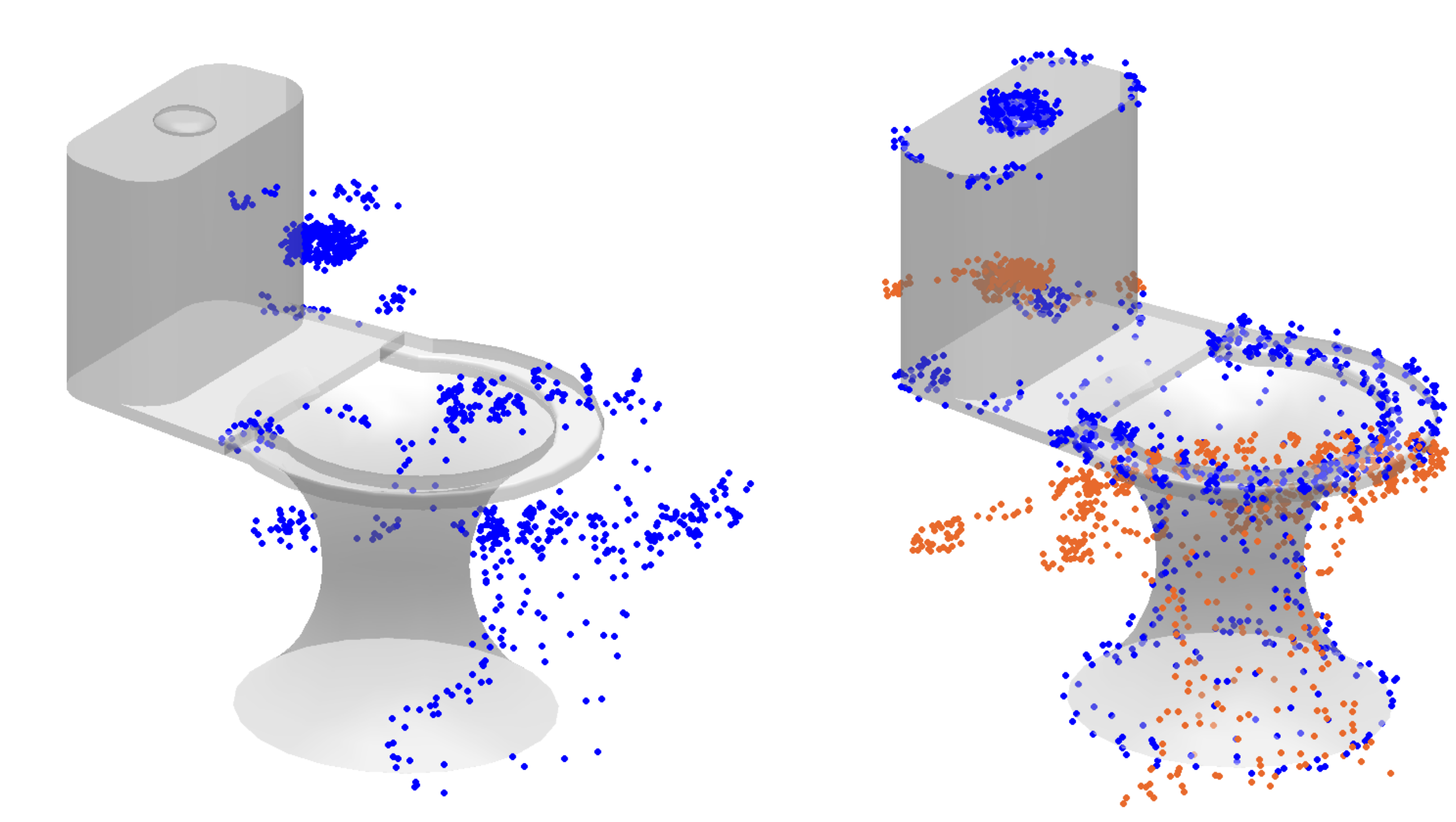}
  \includegraphics[width=0.4\textwidth]{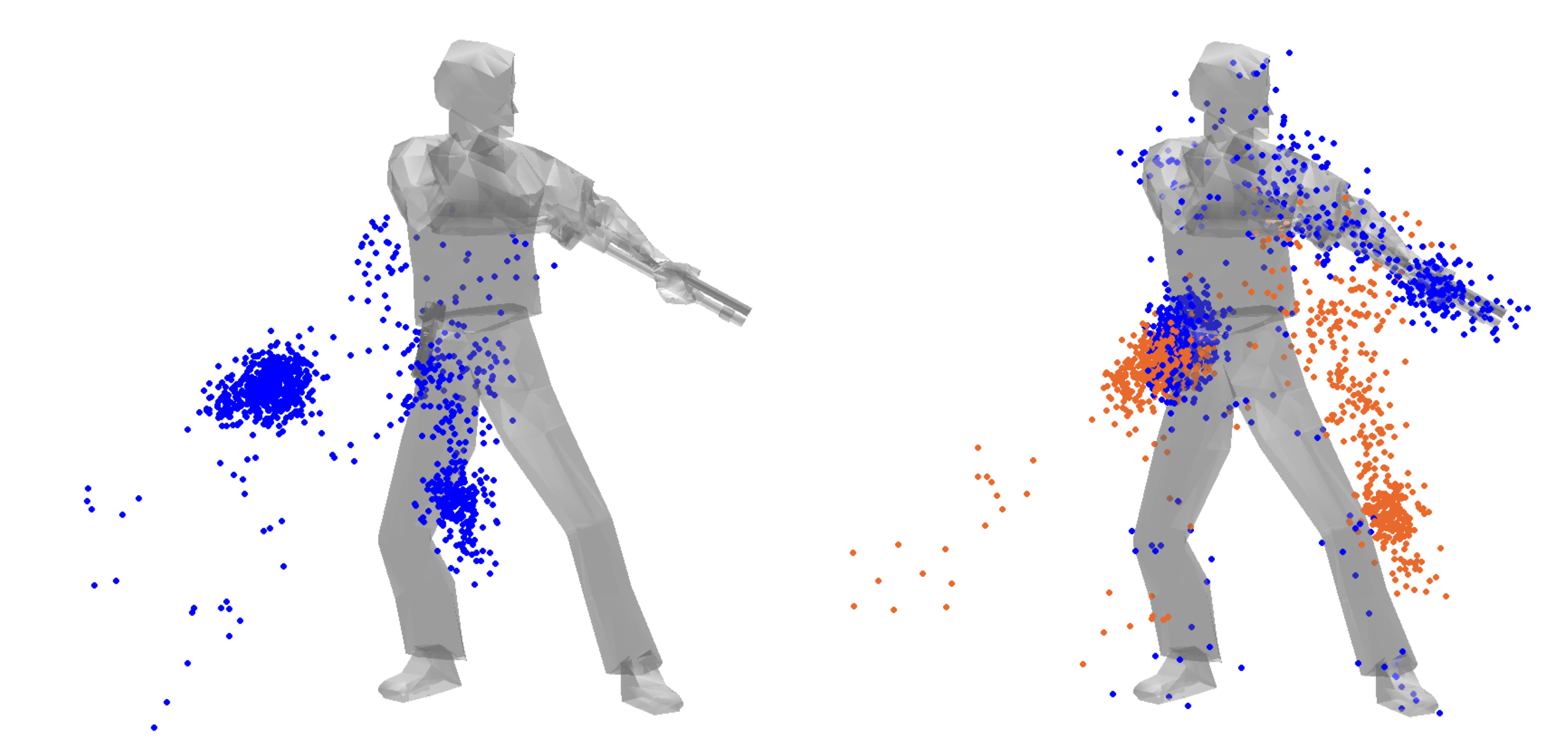}\hspace{10mm}
  \includegraphics[width=0.425\textwidth]{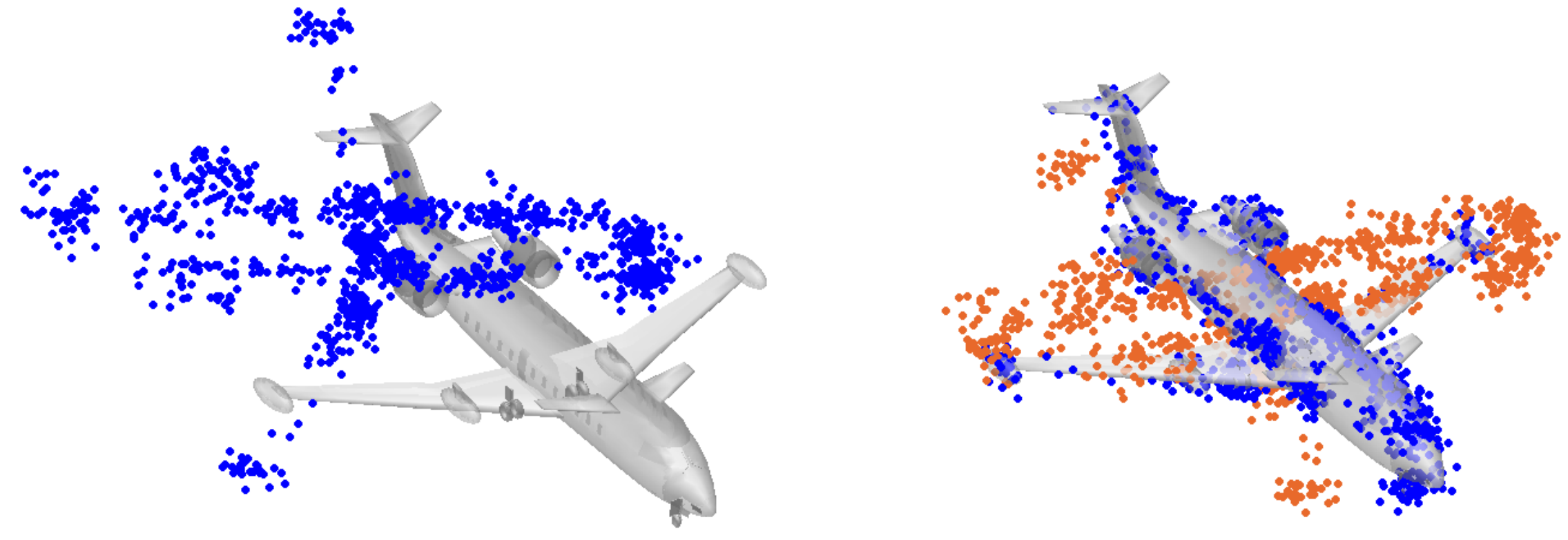}
\caption{Example registrations with Gaussian noise added to each point in the source point cloud for ModelNet object categories unseen during training (Section \ref{gaussian_noise}). For each example, initial position of the points is shown in the left and converged results are shown on the right. The orange points show the ICP estimates and blue points show the PointNetLK estimates.}
\label{fig_gaussiannoise}
\end{figure*}
\subsection{Train and test on same object categories} \label{train_test_same_object_category}

Our first experiment is to train PointNetLK on the training set for 20 object categories in ModelNet40, and test on the test set for the same 20 object categories. We begin by first training a standard PointNet classification network on ModelNet40, and then initialize the PointNetLK feature extractor $\phi$ using this classification network and fine-tune with the PointNetLK loss function. The point clouds used for registration are the vertices from ModelNet40 shapes. The source point cloud is a rigid transformation of the template. Template points are normalized into a unit box at the origin $[0,1]^3$ before warping to create the source. We use random $\textbf{G}_{gt}$ with rotation angles $[0,45]$ degrees about arbitrarily chosen axes and translation $[0,0.8]$ during training of PointNetLK. Results at test time compared with ICP are shown in Fig. \ref{fig_ex1_ex2}. We report results after 10 iterations of both ICP and PointNetLK. This emphasizes an important result, that PointNetLK is able to converge to the correct solution in typically many fewer iterations than ICP. We ensure that testing takes place for the same point clouds and perturbations for both ICP and PointNetLK, for a fair comparison. Initial translations for testing are in the range $[0,0.3]$ and initial rotations are in the range $[0,90]$ degrees.

\subsection{Train and test on different object categories} \label{train_test_different_category}

We repeat the experiment from Section \ref{train_test_same_object_category}, however, we train on the other 20 categories of ModelNet40. We then test on the 20 categories in ModelNet which have not been seen during training, which are the same categories as used in testing for Section \ref{train_test_same_object_category}. We find that PointNetLK has the ability to generalize for accurate alignment on object categories which are unseen during training. The results are shown in Fig. \ref{fig_ex1_ex2} for ModelNet40 test dataset, and Fig.~\ref{fig_bunny} on the Stanford bunny dataset~\cite{turk2005stanford}. The result with Stanford bunny dataset is especially impressive as this dataset is significantly different than the ModelNet training data. For the sake of comparison we also repeated the experiments with ICP and Go-ICP~\cite{Yang13}. We observe that the rotation and translation errors respectively for ICP are $(175.51^\circ,0.22)$, Go-ICP are $(0.18^\circ,10^{-3})$ and PointNetLK are $(0.2^\circ,10^{-4})$. While ICP takes 0.36s, and Go-ICP takes 80.78s, PointNetLK takes only 0.2s.

\subsection{Gaussian noise} \label{gaussian_noise}

We explore the robustness of PointNetLK against Gaussian noise on points. The experiment set-up is as follows: a template point cloud is randomly sampled from the faces of the ModelNet shape, and a source is set equal to the template with additive Gaussian noise of certain standard deviation. We use 1000 points during sampling. We hypothesize that the choice of symmetric operator becomes more critical to the performance of PointNetLK in this experiment. As noted in the original PointNet work, using the max pool operator leads to a critical set of shape points which define the global feature vector. With noisy data, this critical set is subject to larger variation across different random noise samples. Therefore we hypothesize that average pooling would be better suited to learning the global features used for alignment on noisy data. This hypothesis is confirmed in the results shown in Fig. \ref{fig_ex3}.  We repeat the procedure of Section \ref{train_test_different_category}, testing on object categories which are unseen during training. Some example alignment pairs are shown in Fig. \ref{fig_gaussiannoise}.

\subsection{Partially visible data} \label{partially_visible_data}

\begin{figure}[htbp]
\centering
\includegraphics[width=0.4\textwidth]{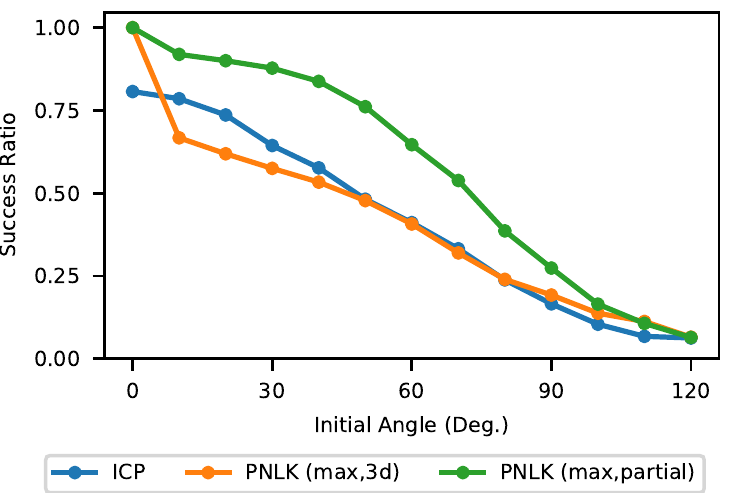}
\caption{Results for Section \ref{partially_visible_data}. We test registration of partially visible ModelNet data, comparing ICP, PointNetLK trained on 3D data, and PointNetLK trained on partially visible data. Both PointNetLK models are trained with max pool. Test categories are unseen during training. We find that training with partially visible data greatly improves performance, even surpassing ICP. A registration is counted as successful if the final alignment rotation error is less than 5 degrees and translation error is less than 0.01. Notice that PointNetLK has perfect performance at zero initial angle since we subtract the mean of each point cloud, whereas ICP does not.}
\label{fig_partiallyvisibleplot}
\end{figure}

We explore the use of PointNetLK on the common registration scenario of aligning 2.5D data. In the real world, oftentimes the template is a full 3D model and the source a 2.5D scan. One approach in this case is to input the 2.5D source and 3D template directly into an alignment algorithm and estimate the correspondence and the alignment. A second approach is to use an initial estimate of camera pose with respect to the 3D model to sample visible points on the model, which can be compared with the 2.5D scan. The camera pose can be iteratively updated until the visible points on the 3D model match the 2.5D scan.

We take the latter approach for testing PointNetLK, because the cost function $\phi(\textbf{P}_\mathcal{T}) - \phi(\textbf{G} \cdot \textbf{P}_\mathcal{S})$ can tend to be large for input point clouds which are a 3D model and 2.5D scan. Instead, it makes more sense to sample visible points from the 3D model first based on an initial pose estimate, so that the inputs to PointNetLK are both 2.5D. This way, a correct final alignment is more likely to lead to the cost function $\phi(\textbf{P}_\mathcal{T}) - \phi(\textbf{G} \cdot \textbf{P}_\mathcal{S})$ being close to zero.

Sampling visible points is typically based on simulating a physical sensor model for 3D point sensing, which has a horizontal and vertical field-of-view, and a minimum and maximum depth~\cite{mehra2010visibility,eckart2018eoe}. We adapt ModelNet40 data for partially visible testing using a simplistic sensor model as follows. We sample faces from ModelNet shapes to create a template, place the template into a unit box $[0,1]^3$, set the template equal to the source, and warp the source using a random perturbation. Next we translate the source and template both by a vector of length $2$ in the direction $[1,1,1]^T$ from the origin. Then we assign the visible points of the template $\textbf{P}_\mathcal{T}^v$ as those satisfying $(\textbf{P}_\mathcal{T} + 2 \cdot [1,1,1]^T) < \text{mean}(\textbf{P}_\mathcal{T} + 2 \cdot [1,1,1]^T)$. This operation can be thought of a placing a sensor at the origin which faces the direction $[1,1,1]^T$ and samples points on the 3D models which lie in front of it, up to a maximum depth equal to the mean of the point cloud. We set the visible source points $\textbf{P}_\mathcal{S}^v$ in the same manner. This operation returns about half of the points both template and source being visible for any given point cloud. We input the 2.5D visible point sets $\textbf{P}_\mathcal{T}^v$ and $\textbf{P}_\mathcal{S}^v$ into PointNetLK, allowing a single iteration to occur for estimation of the aligning transform $\textbf{G}_{est}$. We then warp the original full source model $\textbf{P}_\mathcal{S}$ using the single-iteration guess $\textbf{G}_{est}$, and re-sample $\textbf{P}_\mathcal{S}^v$. We repeat the single-iteration update and visibility re-sampling until convergence. We repeat the same procedure for testing ICP.

\begin{figure}[htbp]
\centering
\includegraphics[width=0.35\textwidth]{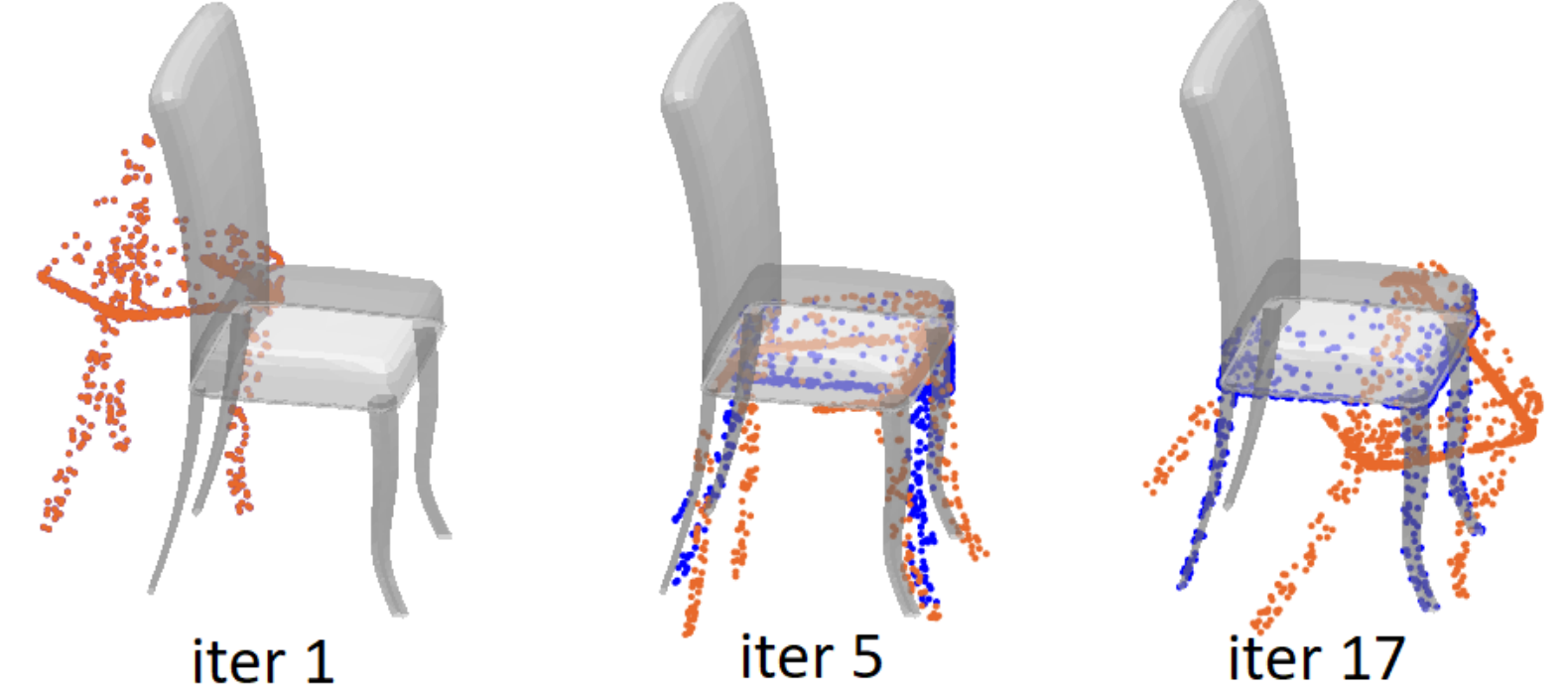}
\hspace{15mm}
\includegraphics[width=0.35\textwidth]{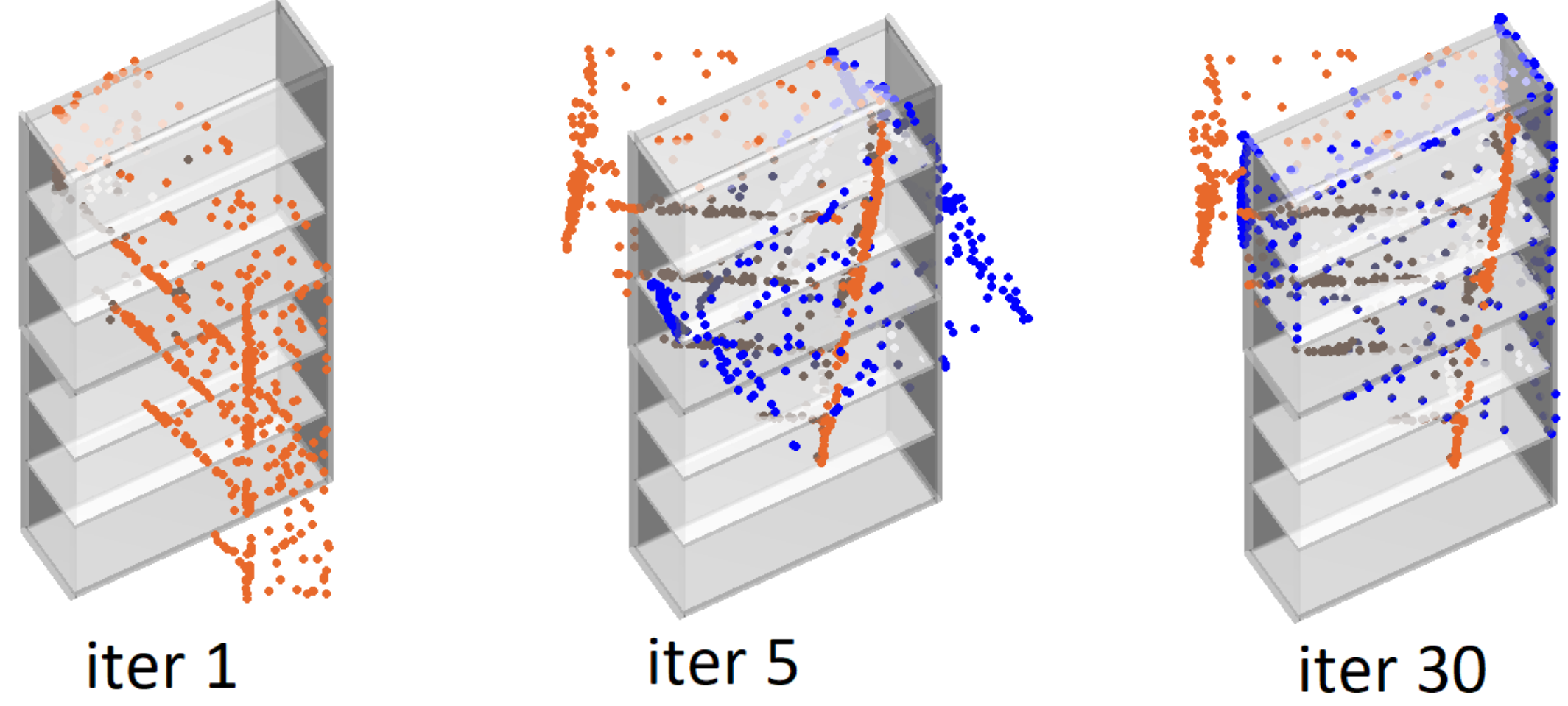}
\includegraphics[width=0.45\textwidth]{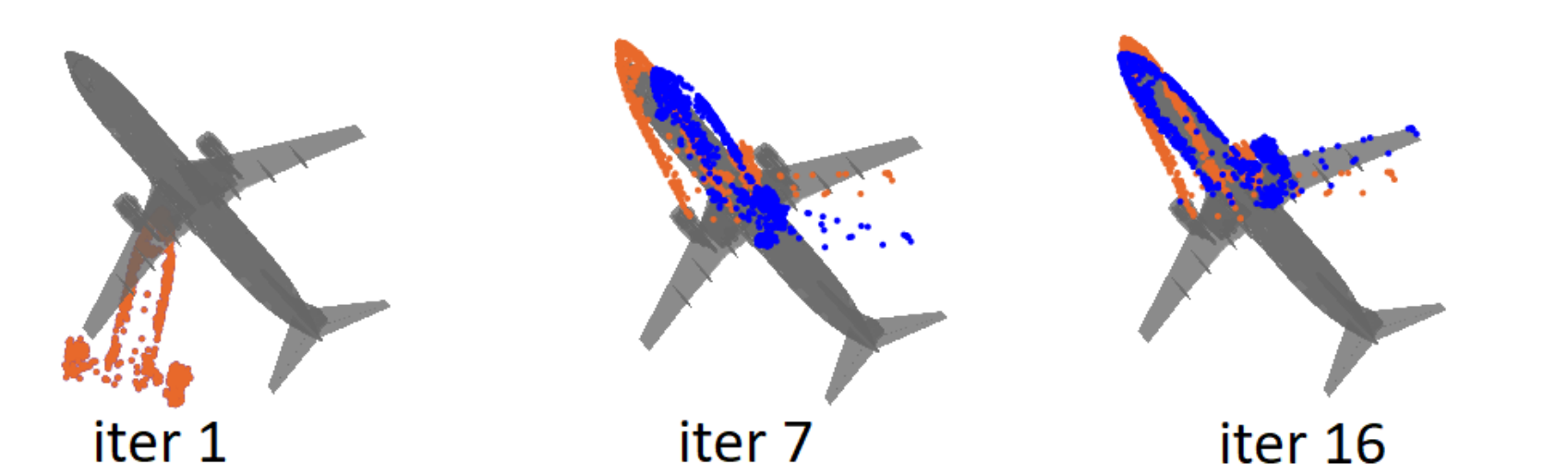}
\includegraphics[width=0.45\textwidth]{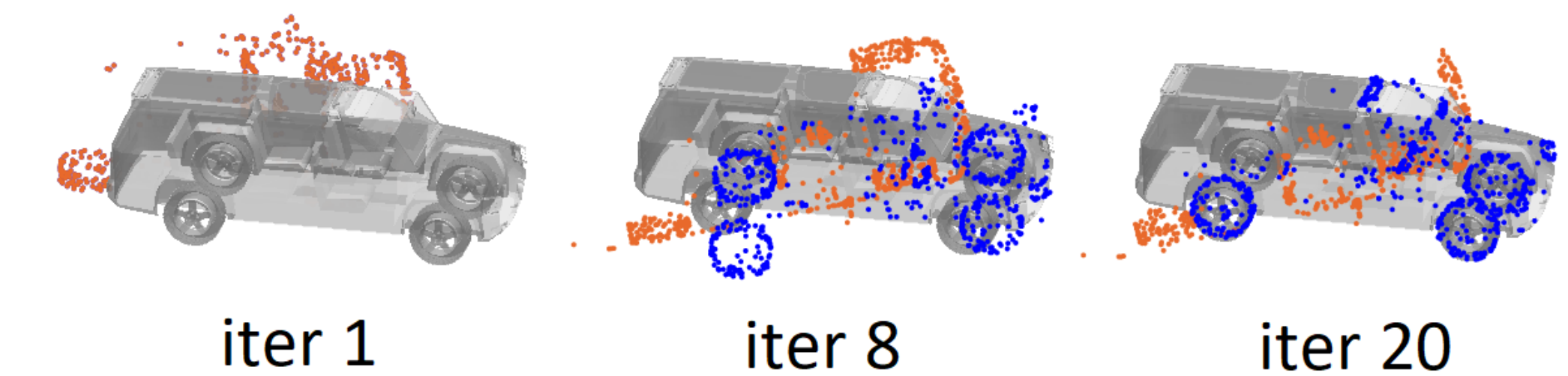}
\caption{Results for Section \ref{partially_visible_data}. We test registration of partially visible ModelNet data, comparing ICP (shown by orange points), and PointNetLK trained on partially visible data (shown by blue points).}
\label{fig_partialregistration}     
\end{figure}

We test on the ModelNet40 test set, using random translation $[0,0.3]$ for all tests. The results are shown in Fig. \ref{fig_partiallyvisibleplot}. Notably, we find that PointNetLK is able to learn to register objects using our sensor model, and generalizes well when the sensor model is applied to unseen object categories. Example template and source pairs for partially visible alignment are shown in Fig. \ref{fig_partialregistration} for ModelNet test dataset. We observe that our approach generalizes well to unseen shapes as shown in Fig.~\ref{fig_bunny} which is generated from RGBD sensor data~\cite{Armeni_2016_CVPR}.

\begin{figure}[htbp]
\centering
\includegraphics[width=0.4\textwidth]{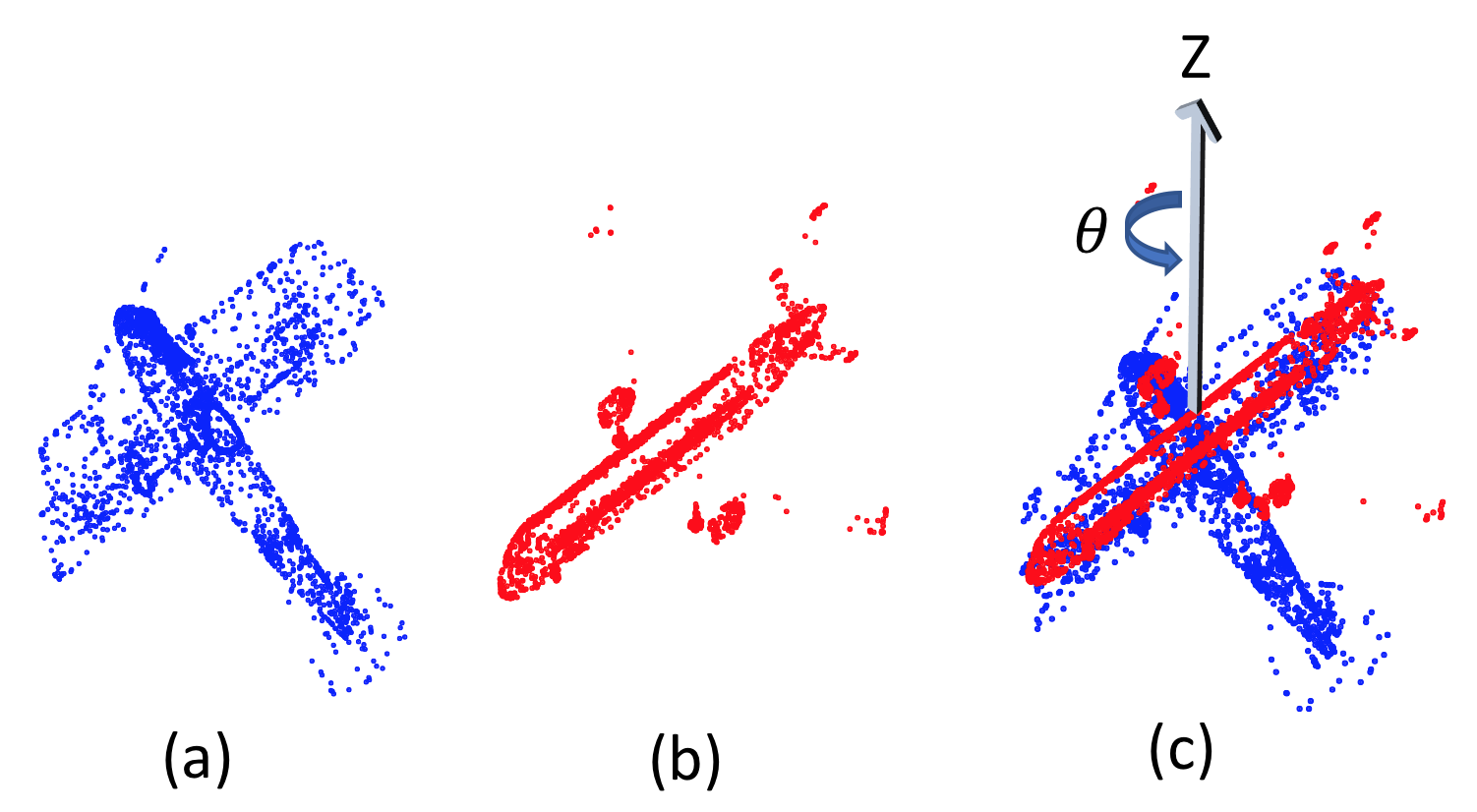}
\includegraphics[width=0.3\textwidth]{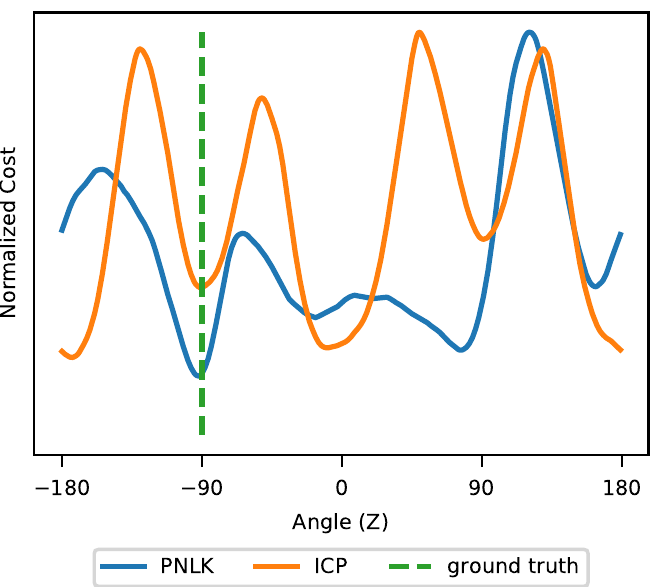}
\caption{Results for Section \ref{same_categ_diff_object}. PointNetLK can achieve a global minimum when two different objects of the same category have the same orientation, whereas ICP can fail. We use two different airplane models from ModelNet40, a biplane (a) and a jetliner (b). (c) shows the initial (incorrect) configuration for alignment, where the centroids each model are at the same location. The jetliner is then rotated about the Z-axis through its centroid. The cost function for standard ICP and PointNetLK during this rotation are plotted. The airplanes have the same orientation at $-90^\circ$ (ground truth). PointNetLK has a global minimum here, whereas ICP has global minimum at $180^\circ$.}
\label{fig_same_categ_diff_object}
\end{figure}

\subsection{Same category, different object} \label{same_categ_diff_object}

We hypothesize that PointNetLK features could be useful for registering point clouds of objects which are different but of the same category. An example of this is shown for two airplane models in Fig.~\ref{fig_same_categ_diff_object}. We would hope that the registration error for PointNetLK $|\phi(\textbf{G} \cdot \textbf{P}_\mathcal{S}) - \phi(\textbf{P}_\mathcal{T})|$ is minimized when the airplane models, despite being different, are aligned in orientation. This reaffirms that the feature vectors learned for alignment are capturing a sense of the object category, and the canonical orientation of that object. The network used for this experiment is trained using max pool on full 3D models. We find that in many cases, such as in the airplane example of Fig.~\ref{fig_same_categ_diff_object}, the PointNetLK cost function is globally minimized when the correct orientation is attained, while the ICP cost function is not necessarily minimized. In practice, this approach could work particularly well to identify the correct orientation of objects within a category if the orientation is known up to one or two axes of rotation.

\subsection{Computational efficiency}

\begin{figure}[htbp]
\centering 
\includegraphics[width=0.40\textwidth]{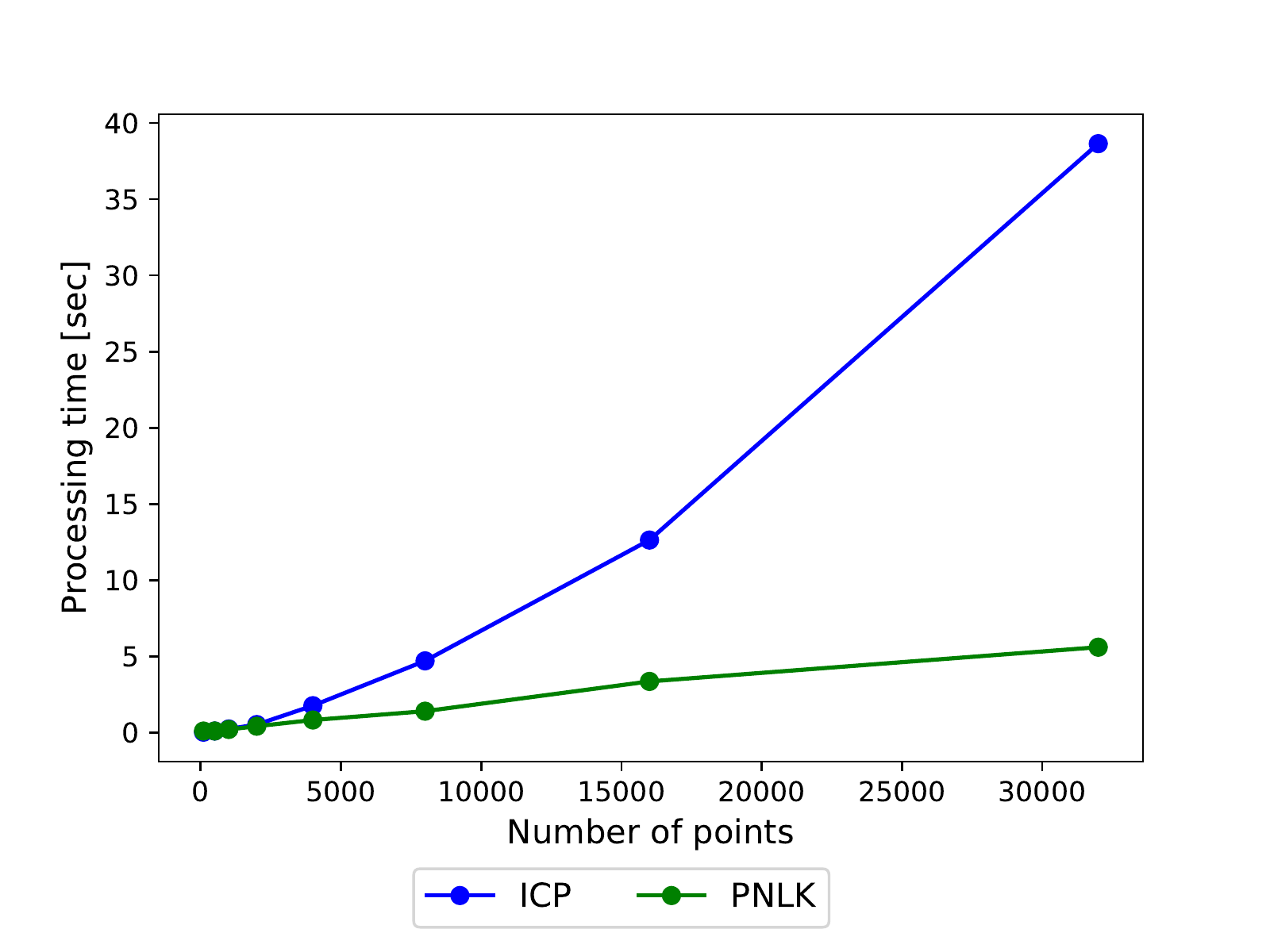}
\caption{Computation cost of PointNetLK grows in $O(n)$ with $n$ points, compared to $O(n^2)$ for ICP.}
\label{fig_NumPoints}
\end{figure}

We plot trends for computation time in Fig. \ref{fig_NumPoints}, comparing PointNetLK and ICP on an Intel Xeon 2GHz CPU. We argue that PointNetLK is quite competitive in efficiency among current approaches to point cloud registration, due to the fact that it has complexity $O(n)$ in $n$ number of points. Note that we do not use a $k$d-tree in the ICP for this particular comparison, because in several applications such as pose tracking from 2.5D data, one does not have $k$d-tree information. Further, the computation can be sped up several orders of magnitude with a GPU implementation as PointNetLK is highly parallelizable.

\section{Implementation Details}

For the MLP in all experiments we use dimensions $(3,64,64,64,128,K=1024)$. Our early experiments showed that this choice of $K$ is suitable for alignment of point clouds containing points on the order of 1000, the number we used in most of our experiments. For setting $t_i$, the infinitesimal perturbations of twist parameters used to compute the Jacobian in Eq.~\ref{eqn_jacob}, we find that $1e^{-2}$ or similar works well. For the minimum threshold for $\Delta \textbf{G}$ used to stop iterations of PointNetLK, we use $|\Delta \xi_i| < 1e^{-7}$. That is, we condition on the magnitude of individual twist parameters which constitute $\Delta \textbf{G}$.

During the fine-tuning stage of training PointNetLK, after training the PointNet classifier, we train for 200 epochs of the ModelNet test set (about one day of training). We find that more epochs are needed to realize good performance for noisy data or partial visibility data (approximately 300 and 400 epochs respectively). When training PointNetLK on 2.5D data, some modifications to the PointNetLK architecture ( as shown in Fig. \ref{fig_block}) were necessary in order to maintain differentiability. This includes creating a visible point mask which sets the non-visible points in the 2.5D source and template to zero, and this mask is applied before the max pooling operator. At test time for 2.5D, differentiability is not a concern and therefore these maskings are not necessary. We implement PointNetLK in PyTorch and train using an \mbox{NVIDIA GeForce GTX Titan X}.

\section{Conclusion} \label{conclusion}
We have presented PointNetLK, a novel approach for adapting PointNet for point cloud registration. We modify the classical LK algorithm to circumvent the inherent inability of the PointNet representation to accommodate gradient estimates through convolution. This modified LK framework is then unrolled as a recurrent neural network from which PointNet is then integrated to form the PointNetLK architecture. Our approach achieves impressive precision, robustness to initialization, and computational efficiency. We have also shown the ability to train PointNetLK on noisy data or partially visible data and achieve large performance gains, while maintaining impressive generalization to shapes far removed from the training set. Finally, we believe that this approach presents an important step forward for the community as it affords an effective strategy for point cloud registration that is differentiable, generalizable, and extendable to other deep learning frameworks.

{\small
\bibliographystyle{ieee}
\bibliography{egbib}
}

\end{document}







\section{Supplementary T-net Experiments}

In the original PointNet work, the authors include transformation networks (T-nets). T-nets are used to regress rigid transformations of points and/or point features which aid with object classification.